\documentclass{article} 
\usepackage{iclr2026_preprint,times}

\usepackage{amsmath,amsfonts,bm}

\def\eqref#1{equation~\ref{#1}}

\def\1{\bm{1}}

\DeclareMathAlphabet{\mathsfit}{\encodingdefault}{\sfdefault}{m}{sl}
\SetMathAlphabet{\mathsfit}{bold}{\encodingdefault}{\sfdefault}{bx}{n}

\PassOptionsToPackage{numbers,compress}{natbib}
\usepackage{graphicx}
\usepackage{amsthm}
\theoremstyle{plain}

\theoremstyle{definition}

\usepackage{wrapfig}
\usepackage{algorithm}
\usepackage{algorithmic}
\usepackage{tabularray}
\usepackage{amssymb}
\usepackage{subcaption}
\usepackage{multirow}
\usepackage{arydshln}
\usepackage{tabularx}
\usepackage[utf8]{inputenc} 
\usepackage[T1]{fontenc}    
\usepackage{hyperref}       
\usepackage{url}            
\usepackage{booktabs}       
\usepackage{amsfonts}       
\usepackage{nicefrac}       
\usepackage{microtype}      
\usepackage{xcolor}         
\usepackage{arydshln}

\title{Sub-Token Routing for KV Cache Compression}

\author{Wei Jiang \\
Futurewei Technologies\\
San Jose, CA 95131, USA \\
\texttt{\{wjiang@futurewei.com} \\
\And
Wei Wang \\
Futurewei Technologies\\
San Jose, CA 95131, USA \\
\texttt{rickweiwang@futurewei.com} \\
}

\iclrfinalcopy 
\begin{document}

\maketitle

\begin{abstract}
Transformer inference often requires a large KV cache, especially for long-context language modeling and multimodal generation. Existing compression methods usually reduce cache cost by selecting, evicting, quantizing, or compressing cached tokens, or by reducing the visual-token sequence before
language-model inference. We introduce sub-token routing, a KV-compression method that adds a finer control axis inside retained tokens. It splits each
retained value vector into groups and keeps only selected groups, while leaving query and key states unchanged. The method is designed to work after token-level reduction. First, a token-reduction method determines which tokens are retained. Then, sub-token routing compresses the value states inside those retained tokens. Experiments under matched KV budgets show that adding sub-token routing improves token-level reduction performance in both LLM and VLM settings,
including Quest on LLaMA-2-7B and Qwen2.5-7B, and FastV/VisionZip across LLaVA and Qwen-VL models. The gains are larger at smaller KV budgets, suggesting that
value-group routing is especially useful when further token removal becomes costly. Overall, token-level reduction and sub-token routing provide complementary ways to reduce KV cost.
\end{abstract}

\section{Introduction}

Transformer inference often requires a large KV cache. This is a common issue
in long-context language modeling and multimodal generation. In LLMs, long
prompts and long generation histories create many cached token states. In VLMs,
each image is converted into many visual tokens, which are processed with the
text tokens and can also be stored in the KV cache. In both cases, the cache
size depends on how many token states are kept and how large each state is.

Most existing methods reduce cache cost by changing the token-level cache or the visual-token sequence. In LLMs, KV-cache methods reduce memory by evicting,
selecting, quantizing, or compressing cached token states
~\cite{kumar2024rvqkv,liu2024minicache,liu2024cachegen,
tang2024quest,xiao2024streamingllm,zhang2023h2o}. Other methods reduce the attention or cache footprint through grouped-query attention, page-level cache
management, head-level allocation, or layer-level reduction
~\cite{ainslie2023gqa,devoto2025expectedattention,shazeer2019fast}. In VLMs, many methods reduce the number of visual tokens given to the language model. FastV prunes visual tokens after early layers~\cite{chen2024fastv}.
LLaVA-PruMerge combines visual-token pruning and merging
~\cite{shang2024prumerge}. SparseVLM and FitPrune use training-free visual-token sparsification based on text relevance or attention preservation
~\cite{ye2024fitprune,zhang2024sparsevlm}. VisionZip selects a smaller set of visual tokens before language-model inference~\cite{yang2024visionzip}. Connector-side methods, such as TokenPacker and Visual Context Compressor, reduce the visual-token sequence before or at the vision--language interface
~\cite{chen2024vcc,li2024tokenpacker}.

These methods use different mechanisms, but most of them choose what to keep at a relatively large unit, such as a token, cache entry, head, page, or layer. After a token is retained, its value representation is usually kept unchanged. We instead consider whether the value representation of a retained token can be
compressed further.

\begin{figure}[t]
\centering
\begin{subfigure}[t]{0.42\textwidth}
    \centering
    \includegraphics[height=4.8cm]{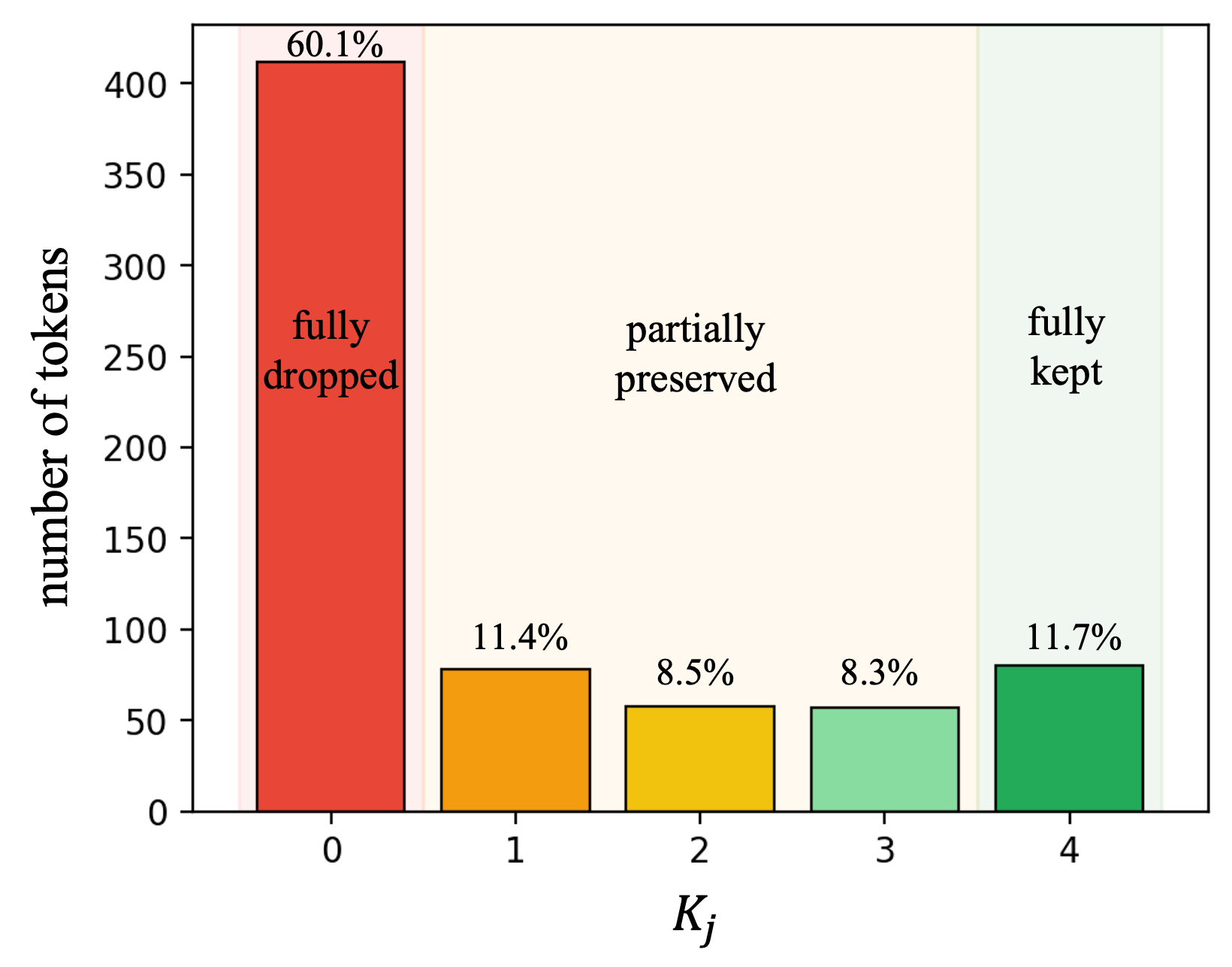}\vspace{-.8em}
    \caption{Histogram of \(K_j\) across tokens.}
    \label{fig:subtoken_diagnostic_a}
\end{subfigure}
\hfill
\begin{subfigure}[t]{0.42\textwidth}
    \centering
    \includegraphics[height=4.8cm]{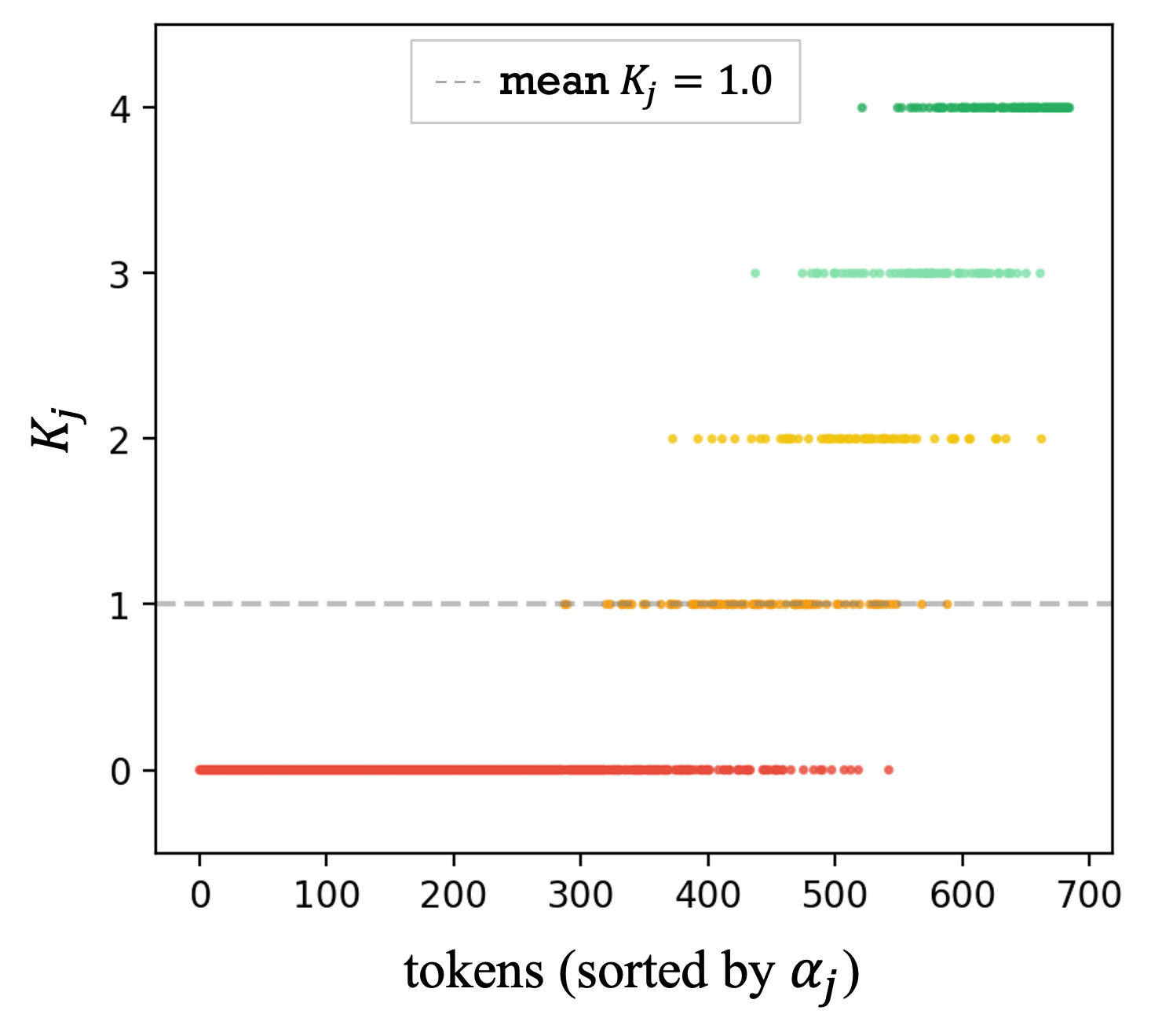}\vspace{-.8em}
    \caption{Token-wise allocation after sorting.}
    \label{fig:subtoken_diagnostic_b}
\end{subfigure}
\caption{
Diagnostics from applying sub-token routing directly in an LLM LoRA setting under a fixed total value-group budget. Here \(K_j \in \{0,\dots,S\}\) denotes
the number of retained value groups for token \(j\), with \(S=4\). (a) The allocation is non-uniform: some tokens receive no value groups, some receive all groups, and many receive a partial allocation. (b) Token-wise allocation after sorting tokens by query-conditioned relevance \(\alpha_j\). Although the average budget is fixed at 1.0, higher-relevance tokens tend to receive larger internal
allocation. The diagnostic suggests that value information is unevenly distributed within token representations, motivating routing below the token
level. }\vspace{-1.5em}
\label{fig:subtoken_diagnostic}
\end{figure}

A token may be useful to the model, but this does not mean that all groups in
its value vector are equally needed. Figure~\ref{fig:subtoken_diagnostic} shows
a diagnostic from applying sub-token routing directly in an LLM LoRA setting.
Under a fixed average budget, the number of retained value groups varies
substantially across tokens. This suggests that value information is not evenly
distributed within token representations.

We introduce \emph{sub-token routing}, which compresses the value path below the
token level. For each token selected for routing, we split the value vector into
groups and retain only selected groups. The query and key states are kept
unchanged, so the attention addressing structure is preserved. Only the value
representation is compressed.

Sub-token routing is designed to work with token-level reduction, not to replace it. A token-level method decides which tokens remain in the sequence or cache.
Sub-token routing then decides how many value groups to keep inside those tokens. The two steps therefore reduce different parts of the KV cache: token selection reduces the number of cached states, while sub-token routing reduces the value width of the retained states.
We study this idea in both LLMs and VLMs. For LLMs, we combine sub-token routing with Quest~\cite{tang2024quest}, a query-aware KV-reduction method. Quest first selects the cached tokens to keep, and sub-token routing then compresses value groups inside those retained tokens. For VLMs, we combine sub-token routing with FastV~\cite{chen2024fastv} and VisionZip~\cite{yang2024visionzip}. The visual-token method first reduces the visual-token sequence, and sub-token routing then compresses the value groups of the remaining visual tokens.

The experiments show that sub-token routing is most useful when token-level compression reaches small budgets. In LLMs, Quest+SubToken improves over Quest on both LLaMA and Qwen models at matched KV budgets, indicating that value-group routing adds useful control beyond query-aware token selection. In VLMs, the gains are larger across both LLaVA and Qwen-VL model families. With FastV and VisionZip, keeping more visual tokens and compressing their value groups preserves more accuracy than pruning to the same visual value-cache budget using token removal alone.

These results show that token-level pruning and sub-token routing are complementary. Token-level pruning reduces the number of retained tokens, while sub-token routing reduces the value states inside those tokens. At small cache budgets, combining the two gives a better accuracy--budget trade-off than using token pruning alone.

Our contributions are as follows:
\begin{itemize}
    \item We formulate sub-token routing for KV-cache compression by applying routing below the token level on the value path.
    \item We keep query and key states unchanged and route only value groups, preserving the attention addressing structure while reducing the retained
    value representation.
    \item We show that sub-token routing is complementary to token-level reduction: it can be applied after token selection to further compress the value states of retained tokens.
    \item We validate this design with matched-budget experiments in both LLMs and VLMs, improving Quest on LLaMA-2-7B and Qwen2.5-7B, and improving FastV
    and VisionZip across LLaVA and Qwen-VL model families.
\end{itemize}

\section{Related Work}

\paragraph{LLM KV-cache compression.}
KV-cache memory grows with context length and decoding history. Existing LLM methods reduce this cost by keeping fewer cached token states or storing them in
a smaller form. H$_2$O keeps heavy-hitter and recent tokens based on attention statistics~\cite{zhang2023h2o}. StreamingLLM uses attention sinks and a sliding-window cache for bounded-memory streaming inference
~\cite{xiao2024streamingllm}. Quest selects cached entries according to their relevance to the current query~\cite{tang2024quest}. Other methods compress or
transmit KV states more compactly, including MiniCache, CacheGen, and RVQ-based KV compression~\cite{kumar2024rvqkv,liu2024minicache,
liu2024cachegen}. Related model and system changes, such as multi-query attention, grouped-query attention, page-level cache management, and head/layer-level allocation, reduce cache or attention cost at larger units
~\cite{ainslie2023gqa,devoto2025expectedattention,shazeer2019fast}. These methods mainly operate at the level of tokens, cache entries, heads, pages, or layers. Our method keeps the token-level decision unchanged and compresses the value representation inside retained tokens.

\paragraph{Visual-token reduction for VLMs.}
Visual tokens are a major source of cost in VLMs because images and videos often produce much longer token sequences than text prompts. Many methods therefore reduce the number of visual tokens before or during language-model inference. FastV prunes visual tokens after early language-model layers using attention
patterns~\cite{chen2024fastv}. LLaVA-PruMerge combines visual-token pruning and merging~\cite{shang2024prumerge}. SparseVLM uses text-guided token scoring without training~\cite{zhang2024sparsevlm}. FitPrune derives pruning decisions by matching attention statistics before and after compression~\cite{ye2024fitprune}. VisionZip selects a smaller set of visual tokens before
language-model inference~\cite{yang2024visionzip}. Other VLM efficiency methods reduce the visual sequence at the vision--language interface or allocate cache
budgets across larger units, including TokenPacker, Visual Context Compressor, Dense Connector, QMoP, MadaKV, and MEDA~\cite{chen2024vcc,li2025madakv,li2024tokenpacker,
li2026qmop,wan2025meda,wang2024denseconnector}. These methods mainly decide which visual tokens, heads, layers, or modalities to keep. Once a visual token is retained, its internal value representation is usually kept in full. Sub-token routing instead compresses value groups inside retained visual
tokens.

\paragraph{Query-aware compression.}
Several compression methods use the current query or prompt to decide what to keep. Quest and expected-attention methods use query information to select or
score cached tokens in LLMs~\cite{devoto2025expectedattention,tang2024quest}.
FastV, SparseVLM, and QMoP use prompt-dependent attention, text-guided scoring, or query-guided projector routing in VLMs~\cite{chen2024fastv,li2026qmop,
zhang2024sparsevlm}. These methods mainly use query information to select tokens, cache entries, or projector paths. Our LLM routing uses the query at a finer level by allocating value-group budget inside retained tokens. For VLMs, we use content-based sub-token routing after visual-token selection,
which keeps the value-group routing separate from the query-aware token selection used by prior VLM reduction methods.

Prior LLM and VLM compression methods mainly reduce the number or size of token-level states. Sub-token routing keeps the token-level decision unchanged and compresses the value groups inside retained tokens. It is therefore
complementary to token-level reduction.

\begin{figure}[t]
\centering
    \includegraphics[width=\textwidth]{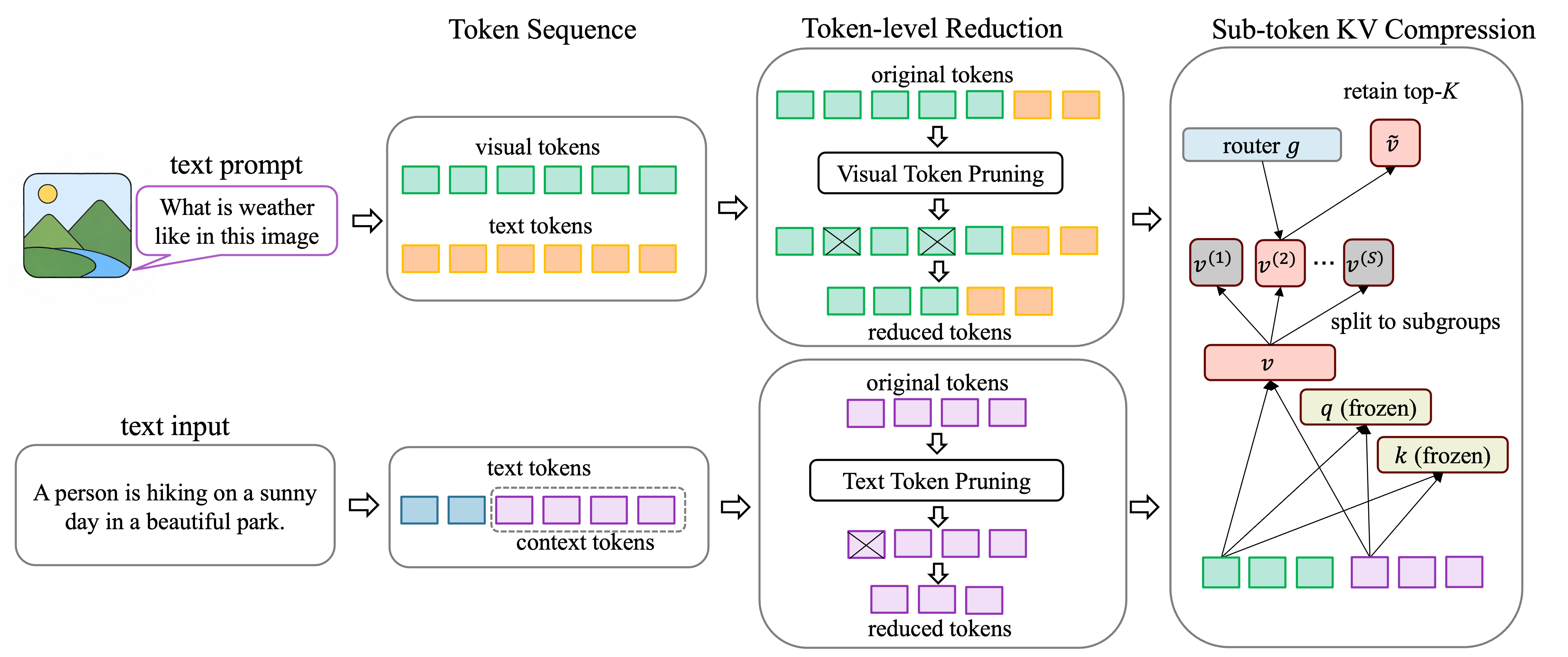}\vspace{-.3em}
\caption{
Overall framework. Token-level pruning first reduces visual tokens in VLMs or context tokens in LLMs. The retained tokens are then processed by sub-token
routing, which keeps query and key states unchanged and selects a subset of value groups inside each retained token.}
\label{fig:method_overview}
\end{figure}

\section{Method}

We study sub-token routing as a value-path compression method for transformer KV caches. The method is applied after token-level reduction. A token-level method first decides which text or visual tokens remain. Sub-token routing then compresses the value representation inside retained tokens.

Figure~\ref{fig:method_overview} gives the overall framework. In LLMs, a token-level KV method selects retained context tokens, and sub-token routing is
applied to the value states of those tokens. In VLMs, a visual-token reduction method first selects retained visual tokens, and sub-token routing is applied to
their value states. In both cases, query and key states are kept unchanged.

\subsection{Preliminaries}

Consider a decoder-only transformer with $L$ layers. Let the input sequence be
\(
X = [x_1,\dots,x_N],
\)
where $N$ is the sequence length after tokenization and optional multimodal
projection. In an LLM, all tokens are text tokens. In a VLM, the sequence
contains \(N_t\) text tokens and \(N_v\) visual tokens:
\[
X = [x_1,\dots,x_{N_t},x_{N_t+1},\dots,x_{N_t+N_v}].
\]
Let \(H^{(\ell)} = [h_1^{(\ell)},\dots,h_N^{(\ell)}]\) denote the hidden states at layer $\ell$. For token $i$, the query, key, and value vectors are computed as
\[
q_i^{(\ell)} = W_q^{(\ell)}h_i^{(\ell)}, \qquad
k_i^{(\ell)} = W_k^{(\ell)}h_i^{(\ell)}, \qquad
v_i^{(\ell)} = W_v^{(\ell)}h_i^{(\ell)} .
\]
During autoregressive inference, keys and values from previous tokens are stored in the KV cache. Token-level methods reduce the number of cached token states.
Our method reduces the value dimension inside the retained states.

\subsection{Stage I: Token-Level Reduction}

We first apply a token-level reduction method. This stage determines
the retained token set. Sub-token routing does not change this token-level
decision.

\paragraph{LLM token-level reduction.}
In the LLM setting, token-level KV reduction operates on context tokens stored
in the cache. Let $\mathcal{C}$ denote this context-token set. A token-level
method assigns each token $j\in\mathcal{C}$ a retention score
\[
u_j = \psi_{\mathrm{tok}}(h_j),
\]
where $\psi_{\mathrm{tok}}$ may use attention statistics, recency, or
query-conditioned relevance. Given a token budget, the method retains the top-ranked tokens according to $u_j$. We denote the retained token set
by $\mathcal{C}_{\mathrm{keep}}$. Sub-token routing is then applied only to tokens
in $\mathcal{C}_{\mathrm{keep}}$.

\paragraph{VLM visual-token reduction.}
In the VLM setting, token-level reduction is applied to visual tokens. Let \(\mathcal{V}=\{1,\dots,N_v\}\) index the visual tokens in the multimodal
sequence. A visual-token reduction method assigns each visual token \(j\in\mathcal{V}\) a score
\[
b_j = \psi_{\mathrm{vis}}(h_j),
\]
which is used to rank visual tokens. Given a token budget, the
method keeps the top-ranked visual tokens according to \(b_j\). We denote the retained visual-token set
by \(\mathcal{V}_{\mathrm{keep}}\). Sub-token routing is then applied only to
tokens in \(\mathcal{V}_{\mathrm{keep}}\).

\subsection{Stage II: Sub-Token Value-Group Routing}
\label{sec:stage2}

The second stage compresses the value path of the retained tokens. The same
operation is used for retained context tokens in LLMs and retained visual tokens
in VLMs.

\paragraph{Value-group decomposition.}
For a retained token \(i\) at layer \(\ell\), let
\(v_i^{(\ell)}\in\mathbb{R}^{d_v}\) denote its value representation. We split it
into \(S\) disjoint groups:
\[
v_i^{(\ell)}
=
\left[
v_i^{(\ell,1)} \mid
v_i^{(\ell,2)} \mid
\cdots \mid
v_i^{(\ell,S)}
\right],
\]
where each group has dimension \(d_v/S\). The routing unit is a value group \(v_i^{(\ell,j)}\).

\paragraph{Routing mask.}
A sub-token router $g^{(\ell)}$ assigns one logit to each value group:
\[
a_i^{(\ell)} = g^{(\ell)}(h_i^{(\ell)}) \in \mathbb{R}^{S}.
\]
For a fixed budget \(K\), the top-ranked $K$ groups are selected, and a binary routing mask $m_{i,j}^{(\ell)}$ is computed, where $m_{i,j}^{(\ell)}=1$ if group $v_i^{(\ell,j)}$ is selected. The routed value groups are therefore
\[
\tilde{v}_i^{(\ell,j)}
=
m_{i,j}^{(\ell)}v_i^{(\ell,j)},
\]
and the routed value vector is
\[
\tilde{v}_i^{(\ell)}
=
\left[
\tilde{v}_i^{(\ell,1)} \mid
\tilde{v}_i^{(\ell,2)} \mid
\cdots \mid
\tilde{v}_i^{(\ell,S)}
\right].
\]
The query and key paths are kept unchanged. Let \(\mathcal{K}^{(\ell)}\) denote the cached key matrix at layer \(\ell\), and let \(\tilde{\mathcal{V}}^{(\ell)}\) denote the routed value matrix. Attention is then computed as:
\[
\mathrm{Attn}\!\left(q_i^{(\ell)},\mathcal{K}^{(\ell)},\tilde{\mathcal{V}}^{(\ell)}\right)
=
\mathrm{softmax}\!\left(
{q_i^{(\ell)}{\mathcal{K}^{(\ell)}}^\top}/{\sqrt{d_h}}
\right)
\tilde{\mathcal{V}}^{(\ell)} .
\]
This keeps the attention addressing structure unchanged and compresses only the
value content.

\paragraph{Compression ratio.}
For a routed token, let \(S\) be the number of value groups and \(K\) be the
number of retained value groups selected by the Top-\(K\) router. The retained value fraction is \(\rho_v = K/S\).  Since key states are kept unchanged and only value states are routed, the
retained KV fraction for a routed token is
\(\rho_{\mathrm{KV}}
=
({1+K/S}){2}
\).  

When sub-token routing is combined with token-level reduction, the token keep
ratio \(r\) further scales the retained cache. The total retained KV fraction is
\[
\rho_{\mathrm{KV}}^{\mathrm{total}}
=
r\cdot \rho_{\mathrm{KV}}.
\]

\subsection{LLM Query-Aware Value-Group Routing}

For LLMs, we use query-aware value-group routing. A cached context token may be
important for one query and less useful for another. We therefore allocate the
value-group budget using information from the query region.

Specifically, the input is divided into a context region \(C\) and a query region \(Q\). The
context region contains cached tokens. The query region contains the tokens used
to determine which cached information is needed. For each retained context token
\(j\in\mathcal{C}_{\mathrm{keep}}\), the router uses the token state and a
query summary to produce group scores:
\[
a_{j}^{(\ell)}
=
g^{(\ell)}\!\left(h_j^{(\ell)},\, s_Q\right),\quad a_j^{(\ell)}\in\mathbb{R}^{S}
\]
where \(s_Q\) is a summary representation of the query region.

Rather than assigning the same number of groups to every retained token, we use
a global token--group budget. Let \(m_{i,j}^{(\ell)}\in\{0,1\}\) denote whether
group \(j\) of token \(i\) is retained. Given a budget \(M_j\), the router keeps
the top-scoring \(M_j\) token--group pairs across
\(\mathcal{C}_{\mathrm{keep}}\):
\[
M_j = \sum\nolimits_{i\in\mathcal{C}_{\mathrm{keep}}}
\sum\nolimits_{j=1}^{S}
m_{i,j}^{(\ell)}.
\]
The number of retained groups can therefore vary by token:
\[
K_i^{(\ell)}
=
\sum\nolimits_{j=1}^{S} m_{i,j}^{(\ell)} .
\]
More relevant context tokens can keep more value groups, while less relevant
tokens keep fewer groups. Query-region tokens are not compressed.

\subsection{VLM Content-Based Value-Group Routing}

For VLMs, we apply sub-token routing after visual-token reduction. Text tokens
are kept unchanged. For each retained visual token
\(i\in\mathcal{V}_{\mathrm{keep}}\), we partition its value representation into
\(S\) groups and apply the value-group mask defined in Sec.~\ref{sec:stage2} to obtain
\(
\tilde{v}_i^{(\ell,j)}\), \(j=1,\dots,S\). 
The routed groups are concatenated to form \(\tilde{v}_i^{(\ell)}\). Query and
key states are unchanged.

In the VLM setting, we use a fixed value-group budget for all routed visual tokens. We also evaluated a query-conditioned visual router that uses the text
prompt, but it gives similar accuracy to content-based routing in our experiments. We therefore use the simpler content-based router in the main VLM results. This keeps visual sub-token routing separate from the visual-token selection method.

\subsection{Training}
\label{sec:training}

We train only the lightweight LoRA adapters and the SubToken routers while keeping the backbone model frozen. Training uses the standard autoregressive
language modeling objective. For VLMs, the input contains the image and text prompt; for LLMs, the input is the text context. Since the router uses discrete top-\(K\) value-group selection, we use a straight-through estimator to backpropagate through the selected groups. Additional training details are given in Appendix~\ref{app:trainingdetails}.

\section{Experiments}
\label{sec:experiments}

\subsection{Experiment Setup}
\label{sec:exp_setup}

\paragraph{Models.}
We evaluate the proposed method in both VLM and LLM settings. For VLMs, we use
LLaVA-1.5-7B, LLaVA-1.5-13B, Qwen2.5-VL-3B, and Qwen2.5-VL-7B. These models
cover both LLaVA-style and Qwen-VL-style architectures and different model
scales. For LLMs, we use LLaMA-2-7B and Qwen2.5-7B to test whether the same
value-group routing idea also improves text-only KV-cache compression.

\paragraph{Training protocol.}
Token-level reduction changes the visual or contextual evidence available to
the language model. We therefore train the LoRA adapter and SubToken router
under the same reduced-token condition used at evaluation. This allows the
\(q/v\) LoRA updates to adapt to the retained token set, while the SubToken
router learns which value groups should be preserved inside those tokens. For
VLMs, the token-level reduction method is FastV~\cite{chen2024fastv} or
VisionZip~\cite{yang2024visionzip}. For LLMs, it is
Quest~\cite{tang2024quest}.

The matched token-only anchor is trained from the same frozen backbone and uses
the same \(q/v\) LoRA capacity, but does not include value-group routing. At
evaluation, the corresponding token-level reduction method is applied to match
the total retained KV budget.

For VLMs, the vision tower is frozen, and SubToken routing is applied only to retained visual tokens; text tokens keep their full KV cache. We train the VLM
LoRA adapters and SubToken routers on the LLaVA-v1.5 instruction-tuning data. We use content-only routing as the default VLM router. For LLMs, SubToken
routing is applied to retained context tokens, and the router is query-conditioned because the importance of a cached token depends on the current query. We train the LLM adapters and routers on WikiText-103  with sequence length \(2048\). In both VLM and LLM settings, the trainable components are limited to the \(q/v\) LoRA adapters and the value-group routers, while the backbone model remains frozen.

\paragraph{Budget accounting.}
SubToken is applied after token-level reduction to retained visual tokens in
VLMs and to retained context tokens in LLMs. All main comparisons use matched
total retained KV fraction. Let \(r\) denote the ratio of kept token by the token-level reduction method. SubToken keeps \(K\) out of \(S\) value groups
inside each retained token. The total
retained KV fraction is
\(
\rho_{\mathrm{KV}}^{\mathrm{total}}
=
r\cdot ({1+K/S}){2}.
\)
The token-only anchor is evaluated at the same retained KV fraction.

\paragraph{Evaluation benchmarks.}
For VLMs, we evaluate on seven benchmarks: MMBench~\cite{liu2023mmbench}, POPE~\cite{li2023pope}, TextVQA~\cite{singh2019textvqa}, MME~\cite{fu2023mme}, GQA~\cite{hudson2019gqa}, VQAv2~\cite{goyal2017vqav2},
and ScienceQA~\cite{lu2022scienceqa}. We report the average over all seven benchmarks as the main aggregate metric. Following commonly reported VLM
token-reduction results~\cite{chen2024fastv,yang2024visionzip}, we also report the 3-benchmark mean over MMBench, TextVQA, and GQA for compact comparison with prior work. For LLMs, we evaluate MMLU 5-shot using all 14{,}042 questions~\cite{hendrycks2021measuring}.

\begin{table*}[t]
\centering
\caption{LLaVA-1.5 results under matched total KV budgets. All SubToken rows use \(K/S=0.25\) after FastV or VisionZip token reduction. FastV25/VisionZip25 combined with SubToken gives 15.625\% total KV, while FastV12.5/VisionZip12.5 combined with SubToken gives 7.8125\% total KV. \(\Delta\) reports the change in 3-mean / 7-mean relative to the matched token-only anchor.}
\label{tab:vlm_llava_all}
\footnotesize
\setlength{\tabcolsep}{1.5pt}
\renewcommand{\arraystretch}{0.92}
\begin{tabular}{lcccccccccc}
\toprule
Method & MMBench & TextVQA & GQA & POPE & MME & SciQA & VQAv2 & 3-m & 7-m & \(\Delta\) 3-m / \(\Delta\) 7-m \\
\midrule
\multicolumn{11}{l}{\textbf{LLaVA-1.5-7B}} \\
\midrule
LoRA-only & 72.00 & 57.87 & 60.45 & 85.24 & 78.18 & 62.42 & 77.03 & 63.44 & 70.46 & -- \\
\addlinespace[0.25em]
\cdashline{1-11}
\addlinespace[0.25em]
\multicolumn{11}{l}{\emph{15.625\% total KV with FastV.}} \\
FastV15.625 & 69.35 & 55.25 & 52.87 & 73.32 & 72.96 & 61.73 & 69.67 & 59.16 & 65.02 & -- \\
FastV25+S4K1 & 69.85 & 55.37 & 56.38 & 79.68 & 74.35 & 62.02 & 71.99 & 61.25 & 67.09 & \(+2.09/+2.07\) \\
FastV25+S8K2 & 70.32 & 55.59 & 55.90 & 79.36 & 74.09 & 61.33 & 72.53 & 61.34 & 67.02 & \(+2.18/+2.00\) \\
FastV25+S16K4 & 69.62 & 56.01 & 56.35 & 80.78 & 75.11 & 62.32 & 72.65 & 61.67 & 67.55 & \(+2.51/+2.53\) \\
\addlinespace[0.25em]
\cdashline{1-11}
\addlinespace[0.25em]
\multicolumn{11}{l}{\emph{15.625\% total KV with VisionZip.}} \\
VisionZip15.625 & 69.67 & 55.09 & 55.82 & 79.66 & 75.15 & 60.78 & 72.83 & 61.25 & 67.00 & -- \\
VisionZip25+S4K1 & 70.52 & 55.28 & 58.08 & 83.26 & 74.85 & 62.77 & 74.78 & 62.71 & 68.51 & \(+1.46/+1.51\) \\
VisionZip25+S8K2 & 70.43 & 54.47 & 57.08 & 83.48 & 76.07 & 60.44 & 72.87 & 61.47 & 67.83 & \(+0.22/+0.83\) \\
VisionZip25+S16K4 & 70.69 & 55.50 & 57.93 & 84.08 & 75.91 & 62.22 & 74.93 & 61.37 & 68.75 & \(+0.12/+1.75\)
\\
\addlinespace[0.25em]
\cdashline{1-11}
\addlinespace[0.25em]
\multicolumn{11}{l}{\emph{7.8125\% total KV with FastV.}} \\
FastV7.8125 & 63.69 & 52.13 & 48.24 & 63.53 & 65.84 & 60.39 & 63.05 & 54.69 & 59.55 & -- \\
FastV12.5+S4K1 & 67.48 & 52.92 & 52.52 & 73.24 & 73.21 & 61.92 & 66.66 & 57.64 & 63.99 & \(+2.95/+4.44\) \\
FastV12.5+S8K2 & 68.51 & 54.65 & 53.35 & 75.30 & 74.43 & 61.77 & 68.36 & 58.79 & 65.20 & \(+4.10/+5.65\) \\
FastV12.5+S16K4 & 68.21 & 54.32 & 54.19 & 78.33 & 73.55 & 61.92 & 69.43 & 59.31 & 65.71 & \(+4.62/+6.16\) \\
\addlinespace[0.25em]
\cdashline{1-11}
\addlinespace[0.25em]
\multicolumn{11}{l}{\emph{7.8125\% total KV with VisionZip.}} \\
VisionZip7.8125 & 66.39 & 52.63 & 51.42 & 75.08 & 71.44 & 61.77 & 67.85 & 56.81 & 63.79 & -- \\
VisionZip12.5+S4K1 & 70.06 & 55.21 & 57.02 & 81.33 & 76.07 & 63.51 & 72.41 & 61.55 & 67.95 & \(+4.74/+4.16\) \\
VisionZip12.5+S8K2 & 69.92 & 55.20 & 57.13 & 81.88 & 76.12 & 62.47 & 72.61 & 60.75 & 67.90 & \(+3.94/+4.11\) \\
VisionZip12.5+S16K4 & 69.58 & 55.28 & 57.02 & 81.73 & 75.78 & 61.92 & 72.85 & 61.72 & 67.74 & \(+4.91/+3.95\) \\
\midrule
\multicolumn{11}{l}{\textbf{LLaVA-1.5-13B}} \\
\midrule
LoRA-only & 74.29 & 61.56 & 61.60 & 87.13 & 76.03 & 66.04 & 78.73 & 65.82 & 72.20 & -- \\
\addlinespace[0.25em]
\cdashline{1-11}
\addlinespace[0.25em]
\multicolumn{11}{l}{\emph{15.625\% total KV with FastV.}} \\
FastV15.625 & 71.06 & 55.27 & 54.78 & 81.82 & 74.01 & 64.75 & 71.91 & 60.37 & 67.66 & -- \\
FastV25+S4K1 & 72.26 & 58.04 & 58.35 & 82.51 & 77.93 & 65.10 & 74.53 & 62.88 & 69.82 & \(+2.51/+2.16\) \\
FastV25+S8K2 & 72.70 & 58.37 & 58.90 & 81.80 & 74.90 & 64.90 & 76.70 & 63.32 & 69.75 & \(+2.95/+2.09\) \\
FastV25+S16K4 & 72.63 & 58.57 & 58.15 & 83.28 & 78.77 & 66.78 & 74.67 & 63.12 & 70.41 & \(+2.75/+2.75\) \\
\addlinespace[0.25em]
\cdashline{1-11}
\addlinespace[0.25em]
\multicolumn{11}{l}{\emph{15.625\% total KV with VisionZip.}} \\
VisionZip15.625 & 71.66 & 56.58 & 48.65 & 71.24 & 65.54 & 64.45 & 64.95 & 58.96 & 63.30 & -- \\
VisionZip25+S4K1 & 71.93 & 58.63 & 59.01 & 84.93 & 78.31 & 65.84 & 76.07 & 63.19 & 70.68 & \(+4.23/+7.38\) \\
VisionZip25+S8K2 & 73.60 & 60.50 & 61.10 & 79.80 & 72.90 & 65.30 & 77.10 & 65.07 & 70.04 & \(+6.11/+6.74\) \\
VisionZip25+S16K4 & 73.20 & 59.07 & 58.90 & 84.14 & 78.39 & 73.00 & 76.53 & 63.72 & 71.89 & \(+4.76/+8.59\) \\
\addlinespace[0.25em]
\cdashline{1-11}
\addlinespace[0.25em]
\multicolumn{11}{l}{\emph{7.8125\% total KV with FastV.}} \\
FastV7.8125 & 65.77 & 49.99 & 47.51 & 74.36 & 68.79 & 65.10 & 61.51 & 54.42 & 61.86 & -- \\
FastV12.5+S4K1 & 71.98 & 56.27 & 56.37 & 81.71 & 75.36 & 65.89 & 72.39 & 61.54 & 68.57 & \(+7.12/+6.71\) \\
FastV12.5+S8K2 & 70.55 & 56.40 & 56.24 & 82.01 & 77.30 & 65.29 & 72.15 & 61.06 & 68.56 & \(+6.64/+6.70\) \\
FastV12.5+S16K4 & 71.08 & 55.81 & 56.23 & 81.52 & 76.62 & 65.69 & 72.55 & 61.04 & 68.50 & \(+6.62/+6.64\) \\
\addlinespace[0.25em]
\cdashline{1-11}
\addlinespace[0.25em]
\multicolumn{11}{l}{\emph{7.8125\% total KV with VisionZip.}} \\
VisionZip7.8125 & 69.79 & 54.53 & 47.06 & 75.14 & 66.93 & 64.80 & 64.58 & 57.13 & 63.26 & -- \\
VisionZip12.5+S4K1 & 72.44 & 58.08 & 58.05 & 83.90 & 78.10 & 66.24 & 74.78 & 62.86 & 70.23 & \(+5.74/+6.97\) \\
VisionZip12.5+S8K2 & 71.73 & 57.97 & 57.65 & 83.88 & 78.43 & 67.13 & 74.20 & 62.45 & 70.14 & \(+5.33/+6.88\) \\
VisionZip12.5+S16K4 & 72.60 & 58.05 & 57.73 & 83.74 & 76.41 & 65.99 & 74.58 & 62.79 & 69.87 & \(+5.67/+6.61\) \\
\bottomrule
\end{tabular}
\vspace{-0.5em}
\end{table*}

\begin{table*}[t]
\centering
\caption{Qwen2.5-VL results under matched total KV budgets. All SubToken rows use $K/S=0.25$ after FastV or VisionZip token reduction. FastV25/VisionZip25 combined with SubToken gives 15.625\% total KV, while FastV12.5/VisionZip12.5 combined with SubToken gives 7.8125\% total KV. $\Delta$ reports the change in 3-mean / 7-mean relative to the matched token-only anchor.}
\label{tab:vlm_qwen_all}
\footnotesize
\setlength{\tabcolsep}{1.5pt}
\renewcommand{\arraystretch}{0.92}
\begin{tabular}{lcccccccccc}
\toprule
Method & MMBench & TextVQA & GQA & POPE & MME & SciQA & VQAv2 & 3-m & 7-m & $\Delta$ 3-m / $\Delta$ 7-m \\
\midrule
\multicolumn{11}{l}{\textbf{Qwen2.5-VL-3B}} \\
\midrule
LoRA-only & 83.11 & 74.74 & 63.41 & 87.60 & 82.90 & 74.42 & 81.27 & 73.75 & 78.21 & -- \\
\addlinespace[0.25em]
\cdashline{1-11}
\addlinespace[0.25em]
\multicolumn{11}{l}{\emph{15.625\% total KV with FastV.}} \\
FastV15.625 & 74.71 & 73.95 & 54.79 & 84.44 & 81.17 & 76.35 & 73.13 & 67.82 & 74.08 & -- \\
FastV25+S4K1 & 78.52 & 71.49 & 59.66 & 86.89 & 80.26 & 73.67 & 76.53 & 69.89 & 75.15 & $+2.07/+1.07$ \\
FastV25+S8K2 & 78.75 & 71.96 & 59.62 & 85.64 & 79.22 & 73.87 & 77.21 & 70.11 & 75.18 & $+2.29/+1.10$ \\
FastV25+S16K4 & 78.49 & 69.25 & 57.18 & 87.26 & 78.23 & 74.57 & 75.19 & 68.31 & 74.31 & $+0.49/+0.23$ \\
\addlinespace[0.25em]
\cdashline{1-11}
\addlinespace[0.25em]
\multicolumn{11}{l}{\emph{15.625\% total KV with VisionZip.}} \\
VisionZip15.625 & 69.74 & 73.31 & 54.44 & 76.72 & 79.95 & 75.06 & 69.28 & 65.83 & 71.21 & -- \\
VisionZip25+S4K1 & 75.88 & 70.01 & 58.15 & 85.41 & 80.50 & 74.52 & 74.71 & 68.01 & 74.17 & $+2.18/+2.96$ \\
VisionZip25+S8K2 & 75.84 & 70.90 & 57.70 & 85.82 & 79.99 & 73.57 & 74.30 & 68.15 & 74.02 & $+2.32/+2.81$ \\
VisionZip25+S16K4 & 75.74 & 70.05 & 57.93 & 85.04 & 80.79 & 74.42 & 75.17 & 67.91 & 74.16 & $+2.08/+2.95$ \\
\addlinespace[0.25em]
\cdashline{1-11}
\addlinespace[0.25em]
\multicolumn{11}{l}{\emph{7.8125\% total KV with FastV.}} \\
FastV7.8125 & 63.18 & 68.53 & 48.49 & 76.84 & 76.20 & 74.32 & 65.63 & 60.07 & 67.60 & -- \\
FastV12.5+S4K1 & 72.44 & 67.71 & 54.42 & 84.29 & 78.37 & 74.37 & 71.39 & 64.86 & 71.86 & $+4.79/+4.26$ \\
FastV12.5+S8K2 & 74.06 & 69.60 & 55.84 & 84.42 & 78.35 & 74.07 & 72.32 & 66.50 & 72.67 & $+6.43/+5.07$ \\
FastV12.5+S16K4 & 73.10 & 69.20 & 55.50 & 83.80 & 75.80 & 74.20 & 72.80 & 65.93 & 72.06 & $+5.86/+4.46$ \\
\addlinespace[0.25em]
\cdashline{1-11}
\addlinespace[0.25em]
\multicolumn{11}{l}{\emph{7.8125\% total KV with VisionZip.}} \\
VisionZip7.8125 & 59.39 & 68.38 & 48.70 & 67.16 & 74.81 & 74.91 & 61.77 & 58.82 & 65.02 & -- \\
VisionZip12.5+S4K1 & 70.29 & 69.14 & 54.96 & 82.52 & 79.23 & 72.58 & 70.40 & 64.80 & 71.30 & $+5.98/+6.28$ \\
VisionZip12.5+S8K2 & 70.25 & 67.37 & 54.69 & 81.93 & 78.01 & 71.89 & 70.31 & 64.10 & 70.64 & $+5.28/+5.62$ \\
VisionZip12.5+S16K4 & 70.80 & 68.09 & 54.42 & 82.71 & 78.01 & 74.12 & 69.97 & 64.44 & 71.16 & $+5.62/+6.14$ \\
\midrule
\multicolumn{11}{l}{\textbf{Qwen2.5-VL-7B}} \\
\midrule
LoRA-only & 86.51 &  78.61 & 63.45 & 87.68 & 82.86 & 83.89 & 81.45 & 74.50 & 80.63 & -- \\
\addlinespace[0.25em]
\cdashline{1-11}
\addlinespace[0.25em]
\multicolumn{11}{l}{\emph{15.625\% total KV with FastV.}} \\
FastV15.625 & 74.77 & 77.93 & 54.14 & 81.31 & 78.06 & 75.90 & 73.74 & 68.95 & 73.69 & -- \\
FastV25+S4K1 & 81.22 & 76.81 & 59.60 & 86.27 & 79.02 & 77.29 & 78.29 & 72.54 & 76.93 & $+3.59/+3.24$ \\
FastV25+S8K2 & 82.12 & 77.16 & 59.34 & 86.69 & 78.55 & 76.40 & 77.57 & 72.87 & 76.83 & $+3.92/+3.14$ \\
FastV25+S16K4 & 82.44 & 78.12 & 59.66 & 86.54 & 78.78 & 77.79 & 78.99 & 73.41 & 77.48 & $+4.46/+3.79$ \\
\addlinespace[0.25em]
\cdashline{1-11}
\addlinespace[0.25em]
\multicolumn{11}{l}{\emph{15.625\% total KV with VisionZip.}} \\
VisionZip15.625 & 72.33 & 75.08 & 52.68 & 78.39 & 75.19 & 78.19 & 69.36 & 66.70 & 71.60 & -- \\
VisionZip25+S4K1 & 79.72 & 75.82 & 58.65 & 86.26 & 79.61 & 77.99 & 76.86 & 71.40 & 76.41 & $+4.70/+4.81$ \\
VisionZip25+S8K2 & 79.83 & 75.65 & 58.50 & 85.56 & 80.03 & 77.59 & 77.27 & 71.33 & 76.35 & $+4.63/+4.75$ \\
VisionZip25+S16K4 & 79.79 & 75.94 & 58.59 & 85.82 & 79.70 & 78.04 & 77.09 & 71.44 & 76.42 & $+4.74/+4.82$ \\
\addlinespace[0.25em]
\cdashline{1-11}
\addlinespace[0.25em]
\multicolumn{11}{l}{\emph{7.8125\% total KV with FastV.}} \\
FastV7.8125 & 61.42 & 71.95 & 48.43 & 73.21 & 73.62 & 72.24 & 63.17 & 60.60 & 66.29 & -- \\
FastV12.5+S4K1 & 73.43 & 75.60 & 55.37 & 84.38 & 69.82 & 74.91 & 73.09 & 68.13 & 72.37 & $+7.53/+6.08$ \\
FastV12.5+S8K2 & 77.15 & 75.59 & 54.97 & 83.78 & 76.56 & 75.01 & 74.77 & 69.24 & 73.98 & $+8.64/+7.69$ \\
FastV12.5+S16K4 & 75.58 & 75.30 & 54.75 & 83.97 & 76.37 & 74.27 & 72.72 & 68.54 & 73.28 & $+7.94/+6.99$ \\
\addlinespace[0.25em]
\cdashline{1-11}
\addlinespace[0.25em]
\multicolumn{11}{l}{\emph{7.8125\% total KV with VisionZip.}} \\
VisionZip7.8125 & 60.89 & 67.93 & 46.72 & 70.33 & 69.39 & 75.21 & 60.53 & 58.51 & 64.43 & -- \\
VisionZip12.5+S4K1 & 73.76 & 72.89 & 54.79 & 84.41 & 77.76 & 74.37 & 72.33 & 67.15 & 72.90 & $+8.64/+8.47$ \\
VisionZip12.5+S8K2 & 73.55 & 72.43 & 54.63 & 84.68 & 77.63 & 75.81 & 71.67 & 66.87 & 72.91 & $+8.36/+8.48$ \\
VisionZip12.5+S16K4 & 72.40 & 71.47 & 53.29 & 82.42 & 76.22 & 74.71 & 70.17 & 65.72 & 71.53 & $+7.21/+7.10$ \\
\bottomrule
\end{tabular}
\vspace{-0.5em}
\end{table*}

\subsection{VLM Results}
\label{sec:vlm_results}

We evaluate VLM compression at small KV budgets, where further token removal can discard visual details needed for downstream reasoning. In this setting,
SubToken is applied with \(K/S=0.25\) after visual-token reduction. A visual-token keep ratio of 25\% gives 15.625\% total KV kept, while a keep
ratio of 12.5\% gives 7.8125\% total KV kept. For each budget, the token-only anchor uses the same total KV fraction, allowing us to compare token-only
reduction with token-plus-value-group reduction under a matched KV budget.

Tables~\ref{tab:vlm_llava_all} and~\ref{tab:vlm_qwen_all} report the VLM results. Across LLaVA and Qwen-VL models, SubToken routing improves over the corresponding token-only anchors for both FastV and VisionZip.

On LLaVA-1.5 models, Table~\ref{tab:vlm_llava_all} shows positive gains at both model scales. For LLaVA-1.5-7B, at 15.625\% total KV, SubToken improves over
the matched anchors by up to $+2.51/+2.53$ for FastV and $+1.46/+1.51$ for VisionZip in 3-mean / 7-mean. At 7.8125\% total KV, the gains increase to $+4.62/+6.16$ for FastV and $+4.91/+4.16$ for VisionZip. For LLaVA-1.5-13B, at 15.625\% total KV, the best gains are $+2.95/+2.09$ for FastV and $+6.11/+6.74$ for VisionZip. At 7.8125\% total KV, the gains reach
$+7.12/+6.71$ for FastV and $+5.74/+6.97$ for VisionZip.

The Qwen results in Table~\ref{tab:vlm_qwen_all} show the same pattern across both model scales. On Qwen2.5-VL-3B, the best gains at 15.625\% total KV are $+2.29/+1.10$ for FastV and $+2.32/+2.96$ for VisionZip. At 7.8125\% total KV, they increase to $+6.43/+5.07$ for FastV and $+5.98/+6.28$ for VisionZip. On Qwen2.5-VL-7B, the gains are stronger. At 15.625\% total KV, SubToken improves over the matched anchors by up to $+4.46/+3.79$ for FastV and $+4.74/+4.82$ for VisionZip, while at 7.8125\% total KV the gains reach $+8.64/+7.69$ and $+8.64/+8.48$, respectively.

\begin{table*}[t]
\centering
\caption{LLM results under matched total KV budgets. Quest25 combined with SubToken gives 23.428\% total KV. Quest12.5 combined with SubToken gives 11.719\% total KV. $\Delta$ reports the change in MMLU relative to matched anchor.}
\label{tab:llm_quest_subtoken}
\footnotesize
\setlength{\tabcolsep}{4pt}
\begin{tabular}{lcc @{\hspace{2.5em}} c @{\hspace{2.5em}} lcc}
\toprule
\multicolumn{3}{c}{\textbf{LLaMA-2-7B}} &&
\multicolumn{3}{c}{\textbf{Qwen2.5-7B}} \\
\cmidrule(lr){1-3}\cmidrule(lr){5-7}
Method & MMLU & $\Delta$ &&
Method & MMLU & $\Delta$ \\
\midrule
Full KV & 45.19 & -- &&
Full KV & 72.90  & -- \\

\midrule
\multicolumn{3}{l}{\emph{23.438\% KV kept}} &&
\multicolumn{3}{l}{\emph{23.438\% KV kept}} \\
Quest23.438 & 42.40 & -- &&
Quest23.438 & 69.87 & -- \\
Quest25+S8K7 & 43.95 & $+1.55$ &&
Quest25+S8K7 & 71.61 & $+1.74$ \\
Quest25+S16K14 & 43.68 & $+1.28$ &&
Quest25+S16K14 & 71.66 & $+1.79$ \\
\midrule
\multicolumn{3}{l}{\emph{21.875\% KV kept}} &&
\multicolumn{3}{l}{\emph{21.875\% KV kept}} \\
Quest21.875 & 40.81 & -- &&
Quest21.875 & 67.55 & -- \\
Quest25+S8K6 & 41.88 & $+1.07$ &&
Quest25+S8K6 & 71.26 & $+3.71$ \\
Quest25+S16K12 & 43.01 & $+2.20$ &&
Quest25+S16K12 & 70.54 & $+2.99$ \\
\midrule
\multicolumn{3}{l}{\emph{11.719\% KV kept}} &&
\multicolumn{3}{l}{\emph{11.719\% KV kept}} \\
Quest11.719 & 30.50 & -- &&
Quest11.719 & 43.13 & -- \\
Quest12.5+S8K7 & 32.47 & $+1.97$ &&
Quest12.5+S8K7 & 44.84 & $+1.71$ \\
Quest12.5+S16K14 & 30.80 & $+0.30$ &&
Quest12.5+S16K14 & 45.11 & $+1.98$ \\
\midrule
\multicolumn{3}{l}{\emph{10.938\% KV kept}} &&
\multicolumn{3}{l}{\emph{10.938\% KV kept}} \\
Quest10.938 & 30.50 & -- &&
Quest10.938 & 40.39 & -- \\
Quest12.5+S8K6 & 31.02 & $+0.52$ &&
Quest12.5+S8K6 & 44.79 & $+4.40$ \\
Quest12.5+S16K12 & 31.73 & $+1.23$ &&
Quest12.5+S16K12 & 44.05 & $+3.66$ \\
\bottomrule
\end{tabular}
\end{table*}

Across VLM model families and scales, the improvement is larger at 7.8125\% total KV than at 15.625\% total KV. This suggests that sub-token routing becomes more
useful when the visual-token set is already small. At this point, removing more visual tokens is likely to discard useful visual evidence, while routing value
groups preserves the retained token set and reduces the value states inside those tokens. The consistent gains across FastV and VisionZip further show that
the effect is not tied to a specific token-selection rule.

The results are also stable across group granularities. For a fixed value-retention ratio \(K/S=0.25\), the \(S=4\), \(S=8\), and \(S=16\) settings give comparable performance within each token-reduction method and
budget. This indicates that the method is not sensitive to the exact group count and remains stable across different value-group partitions.

\subsection{LLM Results}
\label{sec:llm_results}

Table~\ref{tab:llm_quest_subtoken} reports the LLM results on MMLU. In this
setting, Quest provides the token-only anchor, and Quest+SubToken applies
value-group routing inside the tokens retained by Quest. The comparison is made
at matched overall KV fractions.

On LLaMA-2-7B, Quest+SubToken improves over Quest in all tested settings. At
21.875\% KV kept, the best gain is $+2.20$ MMLU points, and at 23.438\% KV
kept, the best gain is $+1.55$ points. The improvement remains positive at the
smaller budgets: Quest+SubToken improves by up to $+1.23$ points at 10.938\%
KV kept and by up to $+1.97$ points at 11.719\% KV kept.

The gains are larger on Qwen2.5-7B. At 21.875\% KV kept, Quest+SubToken
improves over Quest by up to $+3.71$ points, and at 23.438\% KV kept, the best
gain is $+1.79$ points. At the smaller budgets, the improvement reaches
$+4.40$ points at 10.938\% KV kept and $+1.98$ points at 11.719\% KV kept.

These results show that SubToken is not limited to visual-token compression. It
also improves text-only KV-cache compression when combined with a query-aware
token-selection method. The gains are positive across both LLM backbones, but
they are generally smaller than the largest VLM gains. This is reasonable
because Quest already selects retained tokens according to the current query,
leaving less room for value-group routing than in VLM prefix compression, where
token pruning can remove spatial visual evidence. Nevertheless, the consistent
improvement over Quest shows that token-level KV selection and value-group
routing are complementary in text-only models as well.

\subsection{Ablation: Two-Stage Training}
\label{sec:two_stage_ablation}

We further evaluate a two-stage training on LLaVA-1.5-7B. In this ablation, the LoRA-adapted VLM backbone and token-level anchor are first trained and fixed.\, on top of which the SubToken router is then trained. The same LoRA-only baseline and matched token-level anchors from Sec.~\ref{sec:vlm_results} are used.

Table~\ref{tab:vlm_llava7b_twostage} shows that pure value routing is less effective when the KV states and output projection are fixed. All two-stage variants remain below the matched token-level anchors. The gap decreases as the value grouping becomes finer, from S4K1 to S8K2 and S16K4. This trend is much stronger than in the jointly trained main results in  Sec.~\ref{sec:vlm_results}, where different S/K choices give similar performance. This contrast suggests that when the model is jointly optimized, the adapter can absorb much of the difference between value-group configurations. When the token-level anchor is fixed, the router becomes more sensitive to the granularity of the value groups. 

We further test whether the remaining gap comes from router capacity or from the lack of output adaptation. Replacing the linear router with a two-layer MLP improves the results, showing that a more expressive router helps select value groups more effectively. However, the MLP router is still below the matched token-level anchors in most settings. We then train a small output-projection LoRA together with the value router. This ``\_oLoRA" variant keeps the KV cache ratio unchanged. The retained token set and value-group budget are the same as the corresponding SubToken setting, and the additional LoRA only adapts the attention output projection after value routing.

The ``\_oLoRA" results show that lightweight output adaptation can make SubToken effective. At 15.625\% total KV, ``\_oLoRA" nearly matches the FastV anchor in 3-mean and exceeds it in 7-mean, and it also exceeds the VisionZip anchor in 7-mean. At 7.8125\% total KV, ``\_oLoRA" outperforms both FastV and VisionZip matched anchors, improving them by $+2.65/+3.70$ and $+2.09/+2.09$ in 3-mean / 7-mean, respectively. These results indicate that value routing benefits from adapting the computation that consumes the routed values, and that this adaptation can be localized to a small output-projection LoRA without changing the KV cache budget. The jointly trained main model in Sec.~\ref{sec:vlm_results} remains stronger overall, showing that full adapter-router co-adaptation gives the best performance, while ``\_oLoRA" provides a lighter two-stage adaptation path for an already trained token-level compressed model.

\begin{table*}[t]
\centering
\caption{Two-stage LLaVA-1.5-7B ablation under matched total KV budgets. The same LoRA-only baseline and token-level anchors are used as in Sec.~\ref{sec:vlm_results}. ``\_MLP'' methods replace the linear SubToken value router with a two-layer MLP. ``\_oLoRA'' methods train a small output-projection LoRA together with the value router, which does not change the KV cache ratio but only adapts the attention output projection after value-group routing. All SubToken rows use $K/S=0.25$. $\Delta$ reports the change in 3-mean / 7-mean relative to the matched token-level anchor.}
\label{tab:vlm_llava7b_twostage}
\footnotesize
\setlength{\tabcolsep}{1pt}
\begin{tabular}{lcccccccccc}
\toprule
Method  & MMBench & TextVQA & GQA & POPE & MME & SciQA & VQAv2 & 3-m & 7-m & $\Delta$ 3-m / $\Delta$ 7-m \\
\midrule
LoRA-only & 72.00 & 57.87 & 60.45 & 85.24 & 78.18 & 62.42 & 77.03 & 63.44 & 70.46 & -- \\
\midrule
\multicolumn{11}{l}{\emph{15.625\% total KV with FastV.}} \\
FastV15.625  & 69.35 & 55.25 & 52.87 & 73.32 & 72.96 & 61.73 & 69.67 & 59.16 & 65.02 & -- \\
FastV25+S4K1  & 64.91 & 48.55 & 51.26 & 66.17 & 67.99 & 61.48 & 66.38 & 54.91 & 60.96 & $-4.25/-4.06$ \\
FastV25+S8K2  & 65.28 & 50.53 & 52.10 & 72.67 & 69.71 & 60.73 & 67.81 & 55.97 & 62.69 & $-3.19/-2.33$ \\
FastV25+S16K4  & 65.97 & 51.76 & 52.81 & 75.16 & 71.19 & 62.02 & 68.33 & 56.85 & 63.89 & $-2.31/-1.13$ \\
FastV25+S16K4\_MLP  & 66.34 & 52.52 & 53.28 & 76.08 & 70.47 & 61.13 & 68.57 & 57.38 & 64.06 & $-1.78/-0.96$ \\
FastV25+S16K4\_oLoRA  & 67.94 & 54.36 & 55.07 & 78.26 & 71.19 & 61.03 & 70.79 & 59.12 & 65.52 & $-0.04/+0.50$ \\
\midrule
\multicolumn{11}{l}{\emph{15.625\% total KV with VisionZip.}} \\
VisionZip15.625  & 69.67 & 55.09 & 55.82 & 79.66 & 75.15 & 60.78 & 72.83 & 61.25 & 67.00 & -- \\
VisionZip25+S4K1  & 65.00 & 48.29 & 53.39 & 69.29 & 68.70 & 60.83 & 67.96 & 55.56 & 61.93 & $-5.69/-5.07$ \\
VisionZip25+S8K2  & 65.63 & 49.41 & 54.08 & 76.07 & 69.88 & 59.84 & 68.67 & 56.37 & 63.37 & $-4.88/-3.63$ \\
VisionZip25+S16K4  & 66.41 & 51.43 & 54.45 & 80.31 & 71.44 & 61.58 & 70.15 & 57.43 & 65.11 & $-3.82/-1.89$ \\
VisionZip25+S16K4\_MLP  & 66.64 & 52.21 & 54.53 & 79.71 & 71.19 & 61.38 & 69.87 & 57.79 & 65.08 & $-3.46/-1.92$ \\
VisionZip25+S16K4\_oLoRA & 69.53 & 54.12 & 57.63 & 82.84 & 73.76 & 60.54 & 73.71 & 60.43 & 67.45 & $-0.82/+0.45$ \\
\midrule
\multicolumn{11}{l}{\emph{7.8125\% total KV with FastV.}} \\
FastV7.8125  & 63.69 & 52.13 & 48.24 & 63.53 & 65.84 & 60.39 & 63.05 & 54.69 & 59.55 & -- \\
FastV12.5+S4K1  & 61.70 & 47.47 & 47.30 & 58.32 & 62.64 & 60.59 & 60.65 & 52.16 & 56.95 & $-2.53/-2.60$ \\
FastV12.5+S8K2  & 62.25 & 48.73 & 48.30 & 62.91 & 64.91 & 60.49 & 61.73 & 53.09 & 58.47 & $-1.60/-1.08$ \\
FastV12.5+S16K4  & 62.90 & 50.43 & 48.72 & 63.42 & 65.54 & 60.78 & 62.12 & 54.02 & 59.13 & $-0.67/-0.42$ \\
FastV12.5+S16K4\_MLP & 63.20 & 50.63 & 49.54 & 65.18 & 65.21 & 60.98 & 62.73 & 54.46 & 59.64 & $-0.23/+0.09$ \\
FastV12.5+S16K4\_oLoRA & 66.48 & 52.55 & 52.97 & 73.60 & 68.74 & 60.19 & 68.21 & 57.34 & 63.25 & $+2.65/+3.70$ \\
\midrule
\multicolumn{11}{l}{\emph{7.8125\% total KV with VisionZip.}} \\
VisionZip7.8125  & 66.39 & 52.63 & 51.42 & 75.08 & 71.44 & 61.77 & 67.85 & 56.81 & 63.79 & -- \\
VisionZip12.5+S4K1  & 63.09 & 46.25 & 49.92 & 62.78 & 66.98 & 60.78 & 64.49 & 53.08 & 59.18 & $-3.73/-4.61$ \\
VisionZip12.5+S8K2  & 63.39 & 47.76 & 51.28 & 71.28 & 67.61 & 60.44 & 65.83 & 54.14 & 61.08 & $-2.71/-2.71$ \\
VisionZip12.5+S16K4  & 64.77 & 50.19 & 52.34 & 75.12 & 68.49 & 61.68 & 66.81 & 55.77 & 62.77 & $-1.04/-1.02$ \\
VisionZip12.5+S16K4\_MLP & 64.89 & 50.34 & 52.81 & 75.84 & 68.79 & 61.13 & 67.37 & 56.01 & 63.03 & $-0.76/-0.76$ \\
VisionZip12.5+S16K4\_oLoRA & 67.43 & 53.50 & 55.78 & 80.49 & 71.57 & 60.63 & 71.79 & 58.90 & 65.88 & $+2.09/+2.09$ \\
\bottomrule
\end{tabular}
\end{table*}

\section{Conclusion}

We studied sub-token routing for KV-cache compression below the token level. The main idea is to keep the query and key states unchanged while compressing the value states inside retained tokens. This gives a second compression axis that can be used together with token-level reduction: token-level methods decide
which tokens are retained, while sub-token routing reduces the value width of those retained tokens.

Across both LLM and VLM experiments, adding sub-token routing improves matched-budget token-level reduction baselines. In LLMs, combining sub-token routing with Quest improves MMLU on LLaMA-2-7B and Qwen2.5-7B. In VLMs, combining sub-token routing with FastV and VisionZip improves results across LLaVA and Qwen-VL models, with larger gains at smaller KV budgets. The VLM
results also show stable performance across different group granularities, indicating that the method is not sensitive to the exact value-group partition.

These results show that token-level reduction and sub-token routing are complementary ways to reduce KV cost. When the token budget is already small, further token removal may discard useful context or visual evidence. Compressing the value states of retained tokens provides an alternative way to reduce cache
cost while preserving the retained token set.

\bibliographystyle{plainnat}
\bibliography{ref}

\appendix

\section{Extended Related Work}

\subsection{KV-cache compression for LLMs}

KV-cache memory grows with context length and decoding history. Many LLM methods reduce this cost by keeping fewer cached token states or by storing cached states in a smaller form. H$_2$O keeps heavy-hitter tokens and recent tokens based on attention statistics~\cite{zhang2023h2o}. StreamingLLM uses attention sinks and a sliding-window cache for streaming
inference with bounded memory~\cite{xiao2024streamingllm}. Quest selects cached
entries according to their relevance to the current query~\cite{tang2024quest}. Other methods compress or transmit KV states more compactly, including MiniCache, CacheGen, and RVQ-based KV compression
~\cite{kumar2024rvqkv,liu2024minicache,liu2024cachegen}.

Other work reduces attention or cache cost through model and system changes. Multi-query attention and grouped-query attention reduce the number of KV heads
stored and read during inference~\cite{ainslie2023gqa,shazeer2019fast}.
Page-level cache management organizes KV states into reusable memory blocks. Head-level and layer-level allocation methods show that cache budgets need not
be uniform across model components~\cite{devoto2025expectedattention}.

These methods mainly operate at the level of tokens, cache entries, heads, pages, or layers. Our method uses a different granularity. We keep the token-level decision made by a KV-reduction method, but compress the value representation inside retained tokens. This makes the method complementary to query-aware KV selection rather than a replacement for it.

\subsection{Visual-token reduction for VLMs}

Visual tokens are a major source of cost in VLMs. Compared with text prompts, images and videos often produce much longer token sequences, increasing both
prefill computation and KV-cache memory. Many methods therefore reduce the number of visual tokens before or during language-model inference. FastV prunes
visual tokens after early language-model layers using attention patterns~\cite{chen2024fastv}. LLaVA-PruMerge combines visual-token pruning and merging based on token importance and feature similarity~\cite{shang2024prumerge}. SparseVLM uses text-guided token scoring to select visual tokens without training~\cite{zhang2024sparsevlm}. FitPrune derives a pruning recipe by matching attention statistics before and after compression~\cite{ye2024fitprune}. VisionZip selects a smaller set of visual tokens before language-model
inference~\cite{yang2024visionzip}.

These methods mainly decide which visual tokens to keep, remove, or merge. Once a visual token is retained, its internal KV representation is usually kept in full. Sub-token routing keeps the token-level decision unchanged and compresses the value groups inside the retained visual tokens.

Other VLM efficiency methods reduce the visual sequence at the vision--language interface or allocate multimodal cache budgets across larger units. TokenPacker and Visual Context Compressor reduce the number of visual tokens before they enter the language model~\cite{chen2024vcc,li2024tokenpacker}.
Dense Connector and QMoP study compressed or routed visual projectors
~\cite{li2026qmop,wang2024denseconnector}. MadaKV and MEDA allocate KV-cache budgets across modalities, heads, or layers~\cite{li2025madakv,wan2025meda}.
These methods usually act before the language model or at the level of tokens, heads, layers, or modalities. Sub-token routing instead operates inside the value
states of retained tokens.

\subsection{Query-aware compression}

Several compression methods use the current query or prompt to decide what information to keep. In LLMs, Quest selects cached entries according to their
relevance to the current query~\cite{tang2024quest}, and expected-attention methods estimate the usefulness of cached tokens for query processing
~\cite{devoto2025expectedattention}. In VLMs, FastV uses attention patterns affected by the prompt~\cite{chen2024fastv}, and SparseVLM uses text-guided visual-token scoring~\cite{zhang2024sparsevlm}. QMoP also uses a query-guided router over visual projector branches~\cite{li2026qmop}.

These methods mainly use the query to select tokens, cache entries, or projector paths. Our LLM routing uses the query at a finer level: after token-level cache
selection, it allocates value-group budget inside the retained tokens. For VLMs, we use content-based sub-token routing after visual-token selection, since it
matches the accuracy of query-conditioned routing in our experiments while being simpler. This keeps visual sub-token routing separate from the query-aware
token-selection mechanisms used by existing VLM reduction methods.

Overall, prior LLM and VLM compression methods mainly reduce the number or size of token-level states. Sub-token routing keeps the token-level decision unchanged and compresses the value groups inside retained tokens. It is therefore complementary to token-level reduction.

\section{Method Details}
\label{app:method_details}

\subsection{Value-Group Router}
\label{app:vrouter_arch}

Sub-token routing is implemented by attaching a lightweight value-group router to the value path of each language-model attention layer. The router is a single linear projection from the layer input hidden state to \(S\) group logits. The resulting mask is applied only to the value projection. Query states, key states, positional encoding, attention scores, and output projection are unchanged.

The LLM and VLM settings use the same value-path routing principle, but differ in how the value-group budget is allocated. In the LLM setting, the router is
used after Quest and allocates a fixed global budget over the retained
token--group grid. For a retained sequence of length \(N\) and nominal group
budget \(K\), the router selects \(M=NK\) groups from all \(N\times S\)
token--group candidates. This adaptive allocation allows different retained
tokens to receive different numbers of value groups.

In the VLM setting, the router is used after FastV or VisionZip visual-token reduction. The router applies fixed top-\(K\) group selection independently for each retained visual token. Text tokens keep their full value
states. 

Both settings use a straight-through estimator for the discrete group-selection mask. We also use a small load-balancing auxiliary loss to avoid routing collapse across value groups.

\section{Training Details and Implementation Cost}
\label{app:trainingdetails}

\subsection{LLM}

\paragraph{Training configuration.}
For each LLM backbone, we start from a frozen base model and train LoRA adapters together with the SubToken routers. LoRA is applied to the \(q\)- and
\(v\)-projection matrices of every language-model layer. The router is attached to the value path and uses query-conditioned value-group routing. We train on
WikiText-103 with sequence length \(2048\), split into a
\(1024\)-token context region and a \(1024\)-token query region. The objective is autoregressive cross-entropy with a load-balancing auxiliary loss.

\begin{table}[t]
\centering
\caption{LLM training configuration.}
\label{tab:llm_training_config}
\small
\setlength{\tabcolsep}{5pt}
\begin{tabular}{ll}
\toprule
Item & Setting \\
\midrule
Backbones & LLaMA-2-7B, Qwen2.5-7B \\
LoRA & rank \(32\), scaling \(16\), dropout \(0.05\) \\
Router & Per-layer query-conditioned value-group router \\
Quest chunk/window size & \(16 / 64\) \\
Dataset & WikiText-103 \\
Sequence length & \(2048\) \\
Context/query length & \(1024 / 1024\) \\
Optimizer & AdamW \\
Learning rate & \(3\times 10^{-5}\) \\
Weight decay & \(0.01\) \\
Warmup & \(4000\) steps \\
Schedule & Cosine decay after warmup \\
Auxiliary loss & Load-balance loss, weight \(0.01\) \\
Maximum steps & \(25{,}000\) \\
\bottomrule
\end{tabular}
\end{table}

We evaluate two Quest token keep ratios, \(r=0.25\) and \(r=0.125\), and two
value-retention ratios, \(K/S=0.75\) and \(K/S=0.875\). For each ratio, we use
two group granularities, \(S=8\) and \(S=16\).

\paragraph{Evaluation time.}
We evaluate MMLU over all \(14{,}042\) questions. Quest+SubToken is slower than
bare Quest because it adds router evaluation and value-group masking during the
prefix pass.

\begin{table}[t]
\centering
\caption{Mean evaluation time per checkpoint.}
\label{tab:llm_eval_time}
\small
\setlength{\tabcolsep}{5pt}
\begin{tabular}{lcccc}
\toprule
Method & \multicolumn{2}{c}{LLaMA-2-7B} & \multicolumn{2}{c}{Qwen2.5-7B} \\
\cmidrule(lr){2-3}\cmidrule(lr){4-5}
 & MMLU (s) & Total (min) & MMLU (s) & Total (min) \\
\midrule
No compression & \(1357.7\) & \(22.96\) & \(1292.8\) & \(22.08\) \\
Bare Quest & \(1558.0\) & \(26.31\) & \(1475.5\) & \(24.95\) \\
Quest+SubToken & \(2207.1\) & \(37.16\) & \(2173.8\) & \(36.60\) \\
\bottomrule
\end{tabular}
\end{table}

\paragraph{Peak memory.}
Peak memory is measured on a single H100 NVL with bf16 precision, prefix length \(2048\), and a one-token suffix. On LLaMA-2-7B, no compression uses
\(14.791\) GB peak memory. Bare Quest uses \(13.985\) GB at \(25\%\) token keep and \(13.851\) GB at \(12.5\%\) token keep. Quest+SubToken has slightly higher
measured peak memory than bare Quest at the same token keep ratio because the current implementation materializes temporary routing buffers during the
two-pass evaluation, although the steady-state retained KV cache is smaller.

On Qwen2.5-7B, absolute memory savings are smaller because Qwen2.5-7B uses grouped-query attention. At prefix length \(2048\), the measured Qwen2.5-7B KV
cache is about \(0.76\) GB, compared with about \(1.31\) GB for LLaMA-2-7B. The same KV-budget formula applies, but the absolute cache memory available to save
is smaller.

Overall, the LLM implementation adds router computation and temporary buffers on top of Quest. The extra cost comes from routing retained value states, not from
an additional full model forward. The measured wall-clock and peak-memory numbers reflect the current research implementation, and a fused implementation could reduce the temporary-buffer overhead.

\subsection{VLM}

\paragraph{Training configuration.}
For VLM experiments, we train lightweight LoRA adapters together with the SubToken routers while keeping the backbone model, vision tower, and vision-language projector frozen. LoRA is applied only to the \(q\)- and
\(v\)-projection matrices of the language model. The SubToken router is attached to the value path of each language-model layer and is applied only to visual
tokens. Text tokens keep their full value states. The router is content-based in experiments and does not use query conditioning. 

Training uses the LLaVA-v1.5 instruction-tuning data with maximum sequence length \(2048\). During training, the visual-token pruner is active, the LoRA adapters and SubToken routers are trained together from the original released VLM checkpoint.

\begin{table}[t]
\centering
\caption{VLM training configuration.}
\label{tab:vlm_training_config}
\small
\setlength{\tabcolsep}{5pt}
\begin{tabular}{ll}
\toprule
Item & Setting \\
\midrule
Backbones & LLaVA-1.5-7B, LLaVA-1.5-13B, Qwen2.5-VL-3B, Qwen2.5-VL-7B \\
Trainable modules & LM \(q/v\) LoRA adapters and SubToken routers \\
Frozen modules & Base model, vision tower, vision-language projector \\
LoRA & rank \(16\), scaling \(32\), dropout \(0.05\) \\
Router & Per-layer content-based value-group router \\
Dataset & LLaVA-v1.5 instruction-tuning data \\
Training samples & 665K instruction samples \\
Maximum sequence length & \(2048\) \\
Image preprocessing & Aspect-ratio padding \\
Optimizer & AdamW \\
Learning rate & \(2\times 10^{-4}\) \\
Weight decay & \(0.0\) \\
Warmup & \(1000\) steps \\
Auxiliary loss & Load-balance loss, weight \(0.01\) \\
Maximum steps & \(10,000\)\\
\bottomrule
\end{tabular}
\end{table}

\paragraph{SubToken configuration and budget.}
For VLM experiments, we evaluate two visual-token keep ratios, \(r=0.25\) and \(r=0.125\), using FastV or VisionZip. We use a fixed value-retention ratio \(K/S=0.25\), with three group granularities:
\[
(S,K) \in \{(4,1), (8,2), (16,4)\}.
\]
Since keys are kept unchanged, the total retained KV fraction is
\[
\rho_{\mathrm{KV}}^{\mathrm{total}}
=
r \cdot ({1+K/S})/{2}.
\]
Thus, \(r=0.25\) gives 15.625\% total KV, and \(r=0.125\) gives 7.8125\%
total KV. The matched token-only anchors are evaluated at the same total KV
fraction without SubToken routing.

\paragraph{Evaluation time.}
We evaluate VLM inference on a single H100 NVL with bf16 precision. The workload uses one \(336\times336\) image and a short VQA prompt. Prefill time is measured
separately, and decode time is measured with forced 256-token generation so that all configurations use the same number of decode steps.

Table~\ref{tab:vlm_cost_summary} summarizes the measured cost. Visual-token pruning gives the main reduction in prefill cost by shortening the visual-token sequence before language-model inference. Compared with the full LoRA baseline, the token-only FastV and VisionZip anchors reduce the input length from \(595\) to \(163/91\) tokens on LLaVA models and from \(171\) to \(63/45\) tokens on Qwen2.5-VL models.

SubToken routing adds runtime overhead relative to the matched token-only anchor. The overhead comes from evaluating the per-layer value-group router and
applying the value-group mask. In the current implementation, decode time increases by roughly \(22\%\)--\(56\%\) across different settings.
The overhead is similar across \(S=4\), \(S=8\), and \(S=16\), suggesting that the main cost comes from invoking the routing path at each step rather than from
the exact group granularity.

\begin{table*}[t]
\centering
\caption{VLM inference-cost summary on a single H100 NVL. Prefill and decode overheads are measured relative to the matched token-only FastV or VisionZip anchor at the same visual-token keep ratio. Decode overhead is reported over forced 256-token generation.}
\label{tab:vlm_cost_summary}
\small
\setlength{\tabcolsep}{5pt}
\begin{tabular}{lcccc}
\toprule
Model family & Peak alloc. & Pruned input length & Prefill overhead & Decode overhead \\
\midrule
LLaVA-1.5-7B  & \(14.4\)--\(14.7\) GB & \(163 / 91\) & \(+28.9\%\)--\(+46.1\%\) & \(+35.6\%\)--\(+55.2\%\) \\
LLaVA-1.5-13B & \(27.0\)--\(27.5\) GB & \(163 / 91\) & \(+14.8\%\)--\(+45.2\%\) & \(+22.3\%\)--\(+40.3\%\) \\
Qwen2.5-VL-3B & \(7.6\)--\(8.8\) GB   & \(63 / 45\)  & \(+13.0\%\)--\(+29.7\%\) & \(+34.8\%\)--\(+47.1\%\) \\
Qwen2.5-VL-7B & \(16.7\)--\(17.9\) GB & \(63 / 45\)  & \(+10.5\%\)--\(+41.4\%\) & \(+30.9\%\)--\(+55.6\%\) \\
\bottomrule
\end{tabular}
\end{table*}

\end{document}